\documentclass[runningheads]{llncs}
\usepackage{graphicx}
\usepackage{subfigure}
\usepackage{amsfonts}
\usepackage{booktabs}
\usepackage{tabu}
\usepackage{siunitx}
\usepackage{enumitem}
\usepackage{multirow}
\usepackage{capt-of}

\begin{document}
%===========================================================

\newcommand{\etal}{\textit{et al.}}

\title{BAN: Focusing on Boundary Context \\ for Object Detection} % Replace your paper's title here
\titlerunning{BAN: Focusing on Boundary Context for Object Detection} % Replace an abstracted version of your paper's title here

%===========================================================

\author{Yonghyun Kim\inst{1}\orcidID{0000-0003-0038-7850} \and
	Taewook Kim\inst{1}\orcidID{0000-0002-9798-2105} \and
	Bong-Nam Kang\inst{2}\orcidID{0000-0002-6818-7532} \and
	Jieun Kim\inst{1}\orcidID{0000-0002-5552-4165} \and
	Daijin Kim\inst{1}\orcidID{0000-0002-8046-8521}}

%
%Please include author names in full in the paper, 
%If any authors have names that can be parsed into FirstName LastName in multiple ways, please include the correct parsing, in a comment to the volume editors:
%\index{Lastnames, Firstnames}

\authorrunning{Y. Kim~\etal} % A shorter version of authors' name

%===========================================================

\institute{Department of Computer Science and Engineering, POSTECH, Korea\and
	Department of Creative IT Engineering, POSTECH, Korea\\
	\email{ \{gkyh0805,taewook101,bnkang,rlawldmsk,dkim\}@postech.ac.kr}
}

\maketitle

%===========================================================
\begin{abstract}
	Visual context is one of the important clue for object detection
	and the context information for boundaries of an object is especially valuable.
	We propose a boundary aware network~(BAN) designed to exploit the visual contexts including boundary information and surroundings,
	named boundary context,
	and define three types of the boundary contexts: side, vertex and in/out-boundary context.
	Our BAN consists of 10 sub-networks for the area belonging to the boundary contexts.
	The detection head of BAN is defined as an ensemble of these sub-networks with different contributions depending on the sub-problem of detection.
	To verify our method, we visualize the activation of the sub-networks according to the boundary contexts and
	empirically show that the sub-networks contribute more to the related sub-problem in detection.
	We evaluate our method on PASCAL VOC detection benchmark and MS COCO dataset.
	The proposed method achieves the mean Average Precision (mAP) of 83.4\% on PASCAL VOC and 36.9\% on MS COCO.
	BAN allows the convolution network to provide an additional source of contexts for detection and selectively focus on the more important contexts,
	and it can be generally applied to many other detection methods as well to enhance the accuracy in detection.
\end{abstract}

%===========================================================
%===========================================================
\section{Introduction}

% 1. Object Detection의 중요성에 대해서 설명
Object detection is one of the core problem among computer vision tasks because of its extensiveness of applicable areas, such as robotics, visual surveillance and autonomous safety.
In recent years, there have been outstanding achievements in objects detection by successfully deploying a convolutional neural network~\cite{girshick2015fast,li2016r,lin2017feature,liu2016ssd,redmon2016you,redmon2017yolo9000,ren2015faster}.
Despite its success,
there is still a gap between current state-of-the-art performance and perfectness,
and many challenging problems remain unsolved.

%	However, in the field of computer vision, detecting an object from a single image involves many difficulties.
%	In a single image, objects can have a large variation due to external factors such as an unique appearance of the instance, occlusions and various views.
%	This variation increases the ambiguity of the detection problem by complicating the feature space covered by the detection network.

\begin{figure}[t]
	\includegraphics[width=12cm]{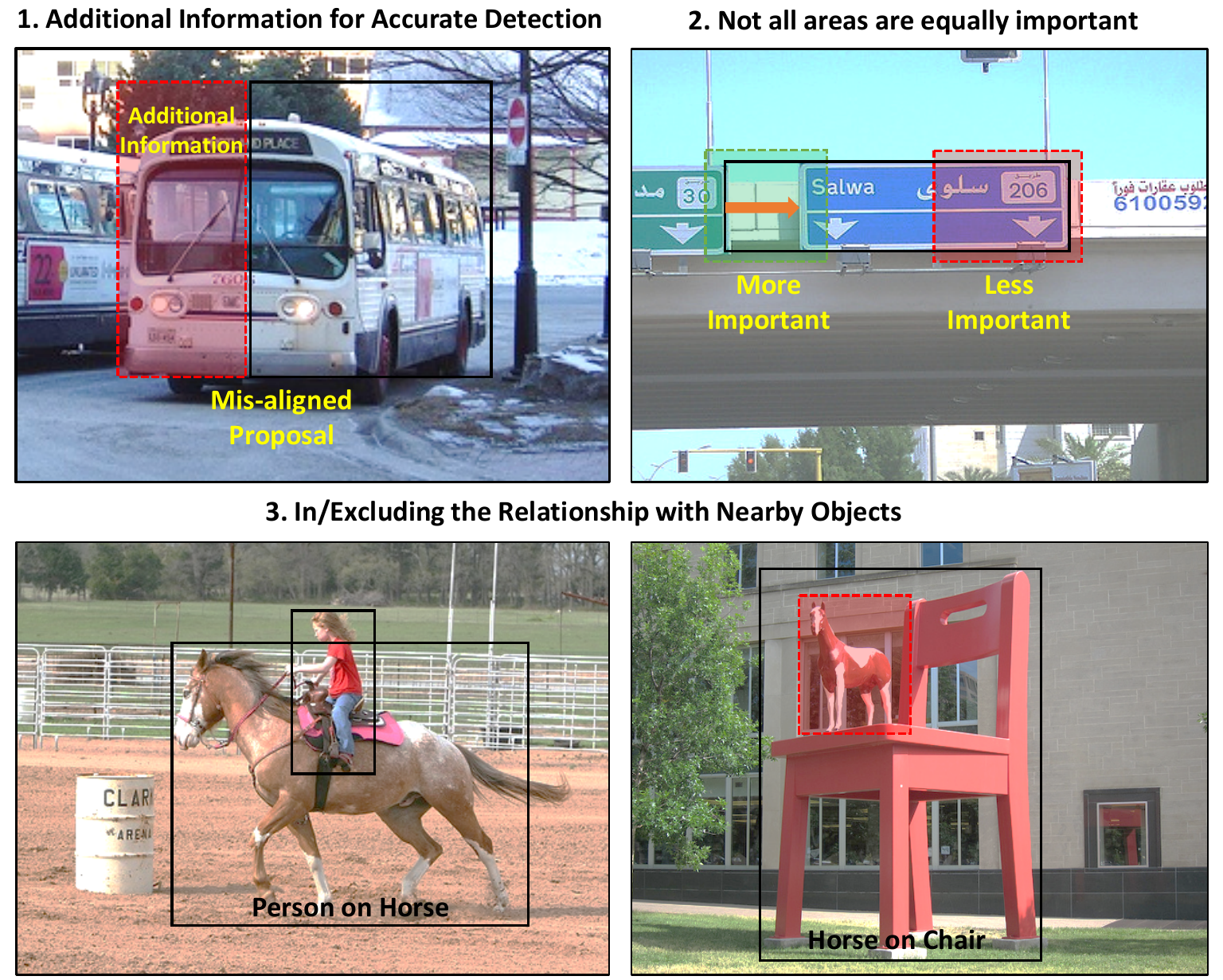}
	\caption{Three advantages of boundary contexts:
		(1) The boundary contexts provide information that could be lost due to mis-aligned proposals for more accurate classification and localization.
		(2) Depending on the sub-problem, the importance of the context may be differently weighted. 
		The detector can localize more accurately by focusing on a specific area.
		(3) As the nearby objects are included or excluded by the context, the relationship between the object of the proposal and the nearby objects can be considered.
		For example, a person on a horse has a valid relationship, but a horse on a chair has a invalid relationship.
	}
	\centering
	\label{fig:advantages}
\end{figure}

Visual context is a powerful clue for object detection
and the context around boundaries of an object such as the surroundings and the shape of the object is especially valuable.
Many advantages can be expected by exploiting the boundary contexts in addition to a given proposal for detection~(Fig.~\ref{fig:advantages}).
The detection frameworks search the objects across the proposals 
generated from region proposal algorithms such as selective search~\cite{uijlings2013selective},
edge boxes~\cite{zitnick2014edge} and region proposal network~\cite{ren2015faster}.
However, mis-aligned proposals with large differences in the location and size of objects may cause
difficulties in detection due to the lack of information.
The boundary context can be an additional source of information for detection and this contexts allow the detector to selectively focus on more important contexts depending on the sub-problem.
The entire network includes and excludes the relationship of the surrounding context,
thereby focusing on the partial detail of the object or considering the relationships between objects.

% 3. 우리는 이러한 Boundary Context를 활용하기 위해 Boundary Aware Network를 제안함
We propose a boundary aware network~(BAN) designed to consider the boundary contexts and empirically prove the effectiveness of BAN.
BAN efficiently represents the relationship between the boundary contexts by implementing the contexts as different sub-networks
and improves the accuracy in detection.
We use a total of 10 boundary contexts from the three different types of pre-defined boundary contexts: side, vertex and in/out-boundary context.
Our BAN consists of 10 corresponding sub-networks for the area belonging to the boundary contexts. 
The detection head of BAN is defined as an ensemble of these sub-networks with different contributions depending on the sub-problem.
We prove the validity of our methods
by visualizing the activation of BAN and measuring the contribution of BAN's sub-networks.

% 실험에 대한 설명 추가해도 됨.
We conduct experiments on two different datasets of object detection and experiments for the strategies for BAN such as a combination of boundary contexts, a feature resolution of sub-networks and sharing of features.
%%% TODO %%%
The proposed BAN shows the improvement of 3.2 mean Average Precision~(mAP) with a threshold of 0.5 IoU from R-FCN~\cite{li2016r} and 1.2 mAP from Deformable R-FCN~\cite{dai2017deformable} on PASCAL VOC~\cite{everingham2010pascal},
and the improvement of 4.5 COCO-style mAP from R-FCN and 2.4 COCO-style mAP from Deformable R-FCN on MS COCO~\cite{lin2014microsoft}.
The experiments verify that BAN improves the accuracy in detection and each boundary context have a distinct meaning for detection.

% 4. 우리의 CONTRIBUTION은 아래와 같다.
We make three main contributions: 
\begin{itemize}[leftmargin=+.4in,label=$\bullet$]
	\item We develop the boundary aware network to consider the boundary contexts around the given proposal 
	and study empirically the influence of the boundary context on classification and bounding box regression.
	Our BAN makes it possible to detect objects more accurately by combining sub-networks of different importance according to the detection head.
	
	\item We empirically demonstrate the effectiveness of BAN for object detection.
	We visualize the activation of the sub-networks according to the boundary contexts and 
	empirically prove that the boundary contexts of BAN contribute more strongly to the detection head if they are intuitively related to each other.
	These related contributions suggest that BAN implies distinct meanings than naive ensemble of sub-networks.
	
	\item 		
	BAN allows the convolution network to provide an additional source of contexts for detection and selectively focus on more important contexts,
	and it can be generally applied to many other detection method as well to enhance the accuracy in detection.
\end{itemize}

% 5. SECTION DETAILS.
This paper is organized as follow.
We review the related works in Section 2.
We demonstrate the proposed BAN and show the effectiveness of BAN in Section 3.
We conduct experiments on two object detection datasets and also present several experiments on the strategies for BAN in Section 4.
We conclude in Section 5.

%------------------------------------------------------------------------- 
\section{Related Works}

\noindent
\textbf{Classic Object Detectors.}
The sliding-window paradigm, in which a classifier is applied on a dense image pyramid~\cite{adelson1984pyramid,gonzalez2009digital}, have been used for a long time to localize objects of various sizes.
Viola and Jones~\cite{viola2004robust} used adaptive boosting with Haar features and a decision stump as a weak classifier primarily for face detection.
Dalal and Triggs~\cite{dalal2005histograms} constructed human detection framework with HOG descriptors and a support vector machine.
Doll\'ar \etal~\cite{dollar2009integral} developed integral channel features, which extract features from channels such as LUV and gradient histogram with integral images, with a boosted decision tree for pedestrian detection.
They expanded it to aggregated channel features and a feature pyramid~\cite{dollar2014fast} for fast and accurate detection framework.
Deformable parts model~(DPM)~\cite{felzenszwalb2010object,yang2013articulated} extend conventional detectors to more general object categories by modelling an object as a set of parts with spatial constraints.
While the sliding-window based approaches had been mainstream for many years,
the advances in deep learning lead CNN-based detectors, described next, to dominate object detection.

\noindent\textbf{Modern Object Detectors.}
The dominant paradigm in modern object detection is a two-stage object detection approach that generates candidate proposals in the first stage and classifies the proposals to the background and foreground classes in the second stage.
The first-stage generators should provide high recall and more efficiency than a sliding window and directly affect the detection accuracy of the second-stage classifiers.
The representative region proposal approaches are selective search~\cite{uijlings2013selective}, edge boxes~\cite{zitnick2014edge} and region proposal network~(RPN)~\cite{ren2015faster}.
As the representative two-stage object detection framework,
Fast and Faster R-CNN~\cite{girshick2015fast,ren2015faster} proposed the standard structure of CNN-based detection and show good accuracy in detection.
These methods extract RoI-wise convolutional features by RoI pooling and classify RoIs of the proposals to the background and foreground classes using RoI-wise sub-networks.  
Region-based fully convolutional networks (R-FCN)~\cite{li2016r} improved speed by designing the structure of networks as fully convolutional by excluding RoI-wise sub-networks.
However, two-stage decision makes the detectors not practical enough.
One-stage detectors such as SSD~\cite{liu2016ssd} and YOLO~\cite{redmon2017yolo9000} showed practical performance by focusing on the speed/accuracy trade-off.
These detectors have a 5-20\% lower accuracy in detection with 30-100 FPS.
We experiment our BAN with R-FCN and show the improvement in the detection accuracy.

\noindent
\textbf{Residual Network.}
The residual network~\cite{he2016deep}, one of the most widely used backbone networks in recent years, was proposed to solve the problem that learning becomes difficult as the network becomes deeper.
Against the expectation that stacking more layers increases accuracy with more capacity,
deeper networks exposed to a degradation of both training and test accuracy.
The degradation of training accuracy implies that the difficulty of learning from deep structures, rather than over-fitting, causes the degradation.
The residual learning prevents the deeper networks from having a higher training error than the shallower networks by adding shortcut connections that are identity mapping.
It is easier for the residual block to learn the residual to zero than to learn the desired mapping function directly.
By designing the desired mapping as a residual function, the residual block makes learning easier for deeper networks.

\noindent
\textbf{Detection with Context.}
Context is an important clue in the applications of computer vision such as detection~\cite{ding2012contextual,divvala2009empirical}, segmentation~\cite{avidan2006spatialboost} and recognition~\cite{carbonetto2004statistical}.
Ding \etal~\cite{ding2012contextual} designed the contextual cues in spatial, scaling and color spaces and developed an iterative classification algorithm called contextual boost.
AZ-Net~\cite{lu2016adaptive} accurately localizes an object by dividing and detecting the region recursively.
Because the divided regions quite differ from the object area at first, it uses the inner and surrounding contexts to iteratively complement the imperfectness of the regions. 
Deformable R-FCN~\cite{dai2017deformable} is a generalization of atrous convolution.
It partially includes the effect of the visual context by exploring the surrounding at the cell level.
FPN~\cite{lin2017feature}/RetinaNet~\cite{lin2017focal} exploit the contexts for scale by aggregating multi-scale convolutional blocks.
These methods try to consider the contextual cues in various ways, however, they partially exploit the visual context. 
BAN provides the distinct context more directly for surroundings and can improve the performance of various detectors easily.

\begin{figure}[t]
	\includegraphics[width=12cm]{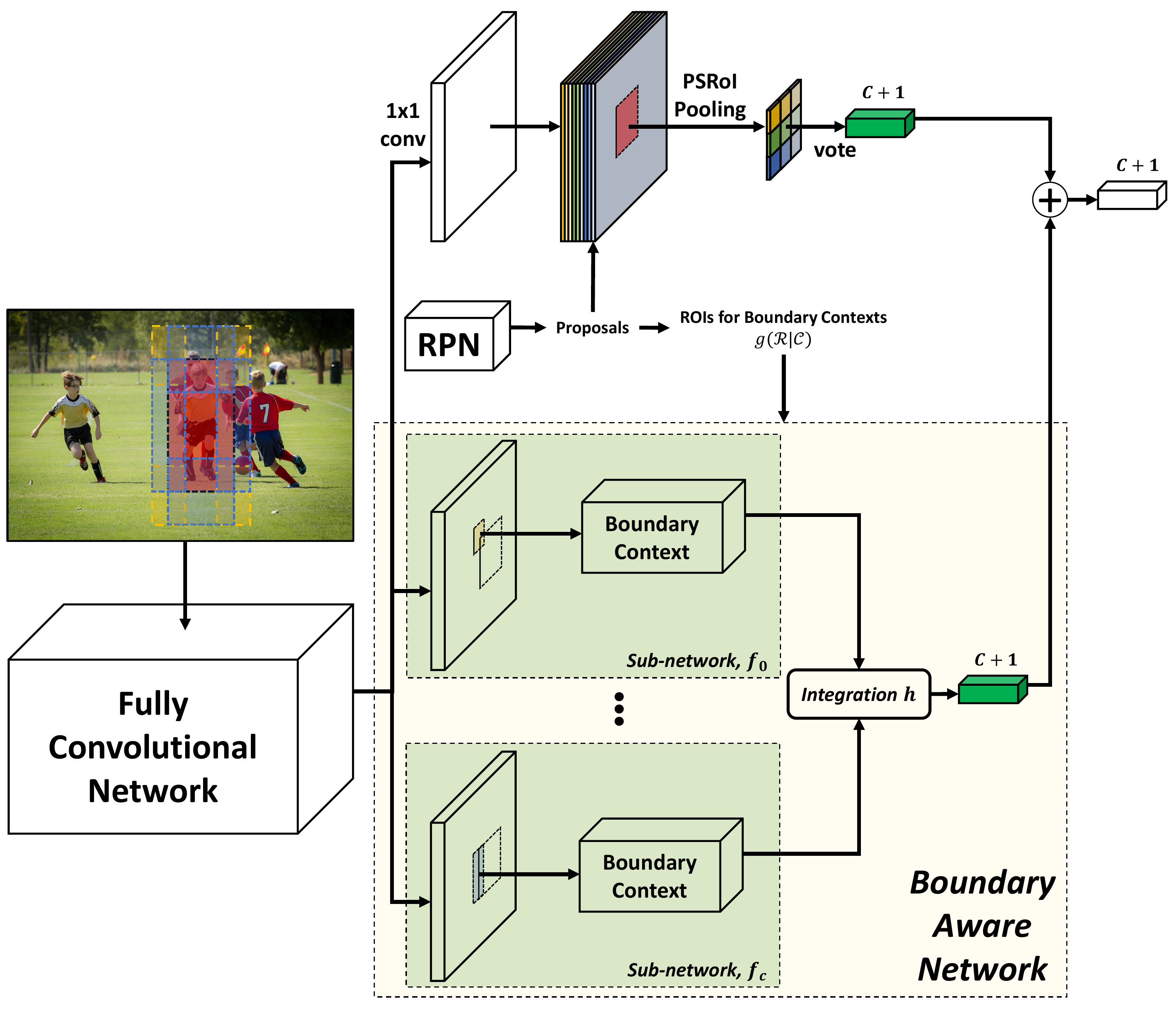}
	\caption{Overview of the proposed BAN with a classifier head for $C$ classes. 
		Our detection architecture classifies and localizes an object from a proposal by integrating sub-networks representing difference boundary contexts.
	}
	\centering
	\label{fig:overview}
\end{figure}
%------------------------------------------------------------------------- 
\section{Boundary Aware Network}

We propose a boundary aware network~(BAN) to exploit the contexts for boundary information and surroundings, named boundary context,
and define three types of the boundary contexts: side, vertex and in/out-boundary context. 
Visual context~\cite{avidan2006spatialboost,carbonetto2004statistical,ding2012contextual,divvala2009empirical} is one of the important clue for object detection.
Because most of the detection frameworks pool convolutional features only from the proposal area, 
it is difficult to directly consider the areas not included exactly in the proposal and the relationship with the surroundings.
The proposed BAN enhance the accuracy in detection 
by ensembling sub-networks that directly use boundary context of the proposal as additional information.

Here, $\mathcal{R}$ is one of the proposals for a given image~$\mathbf{x}$ and $\mathcal{C}$ is a set of the boundary contexts~$c$.
$g(\mathcal{R}|c)$ denotes a generator that provides the boundary region related to $\mathcal{R}$.
The classifier and regressor~$f$, that are the aggregation of detection~$f_0$ for the original proposal and detection~$h$ for BAN that integrates corresponding sub-networks~$f_c$ of each boundary context, are defined in the following form:

\begin{equation}
f(\mathbf{x},\mathcal{R}) = f_{0}(\mathbf{x},\mathcal{R}) + h( \left \{ f_c(\mathbf{x},g(\mathcal{R}|c) | c \in \mathcal{C} \right \} ).
\label{eq:1}
\end{equation}

$h$ is empirically built according to pooling methods such as RoI pooling and PSRoI pooling.
In PSRoI pooling based implementation, each of $f_c$ is a detection head and $h$ is a simple aggregation of the detection heads, and $f$ is defined as the aggregation of baseline and sub-networks of BAN in Eq.~\ref{eq:1}.
Thus, the propagated errors are equally transferred to each sub-network in the back-propagation: 
\begin{equation}
\frac{\partial E}{\partial f} = \frac{\partial E}{\partial f_0} = \frac{\partial E}{\partial f_c}.
\end{equation}
Because the error of the upper layer is propagated equally to each sub-network,
sub-networks are learned in a balanced manner considering the importance of each context for the same goal.
In Section~\ref{section:sub1}, we show that each sub-network of BAN actually contributes more to the related sub-problem.

\begin{figure}[!t]
	\includegraphics[width=12cm]{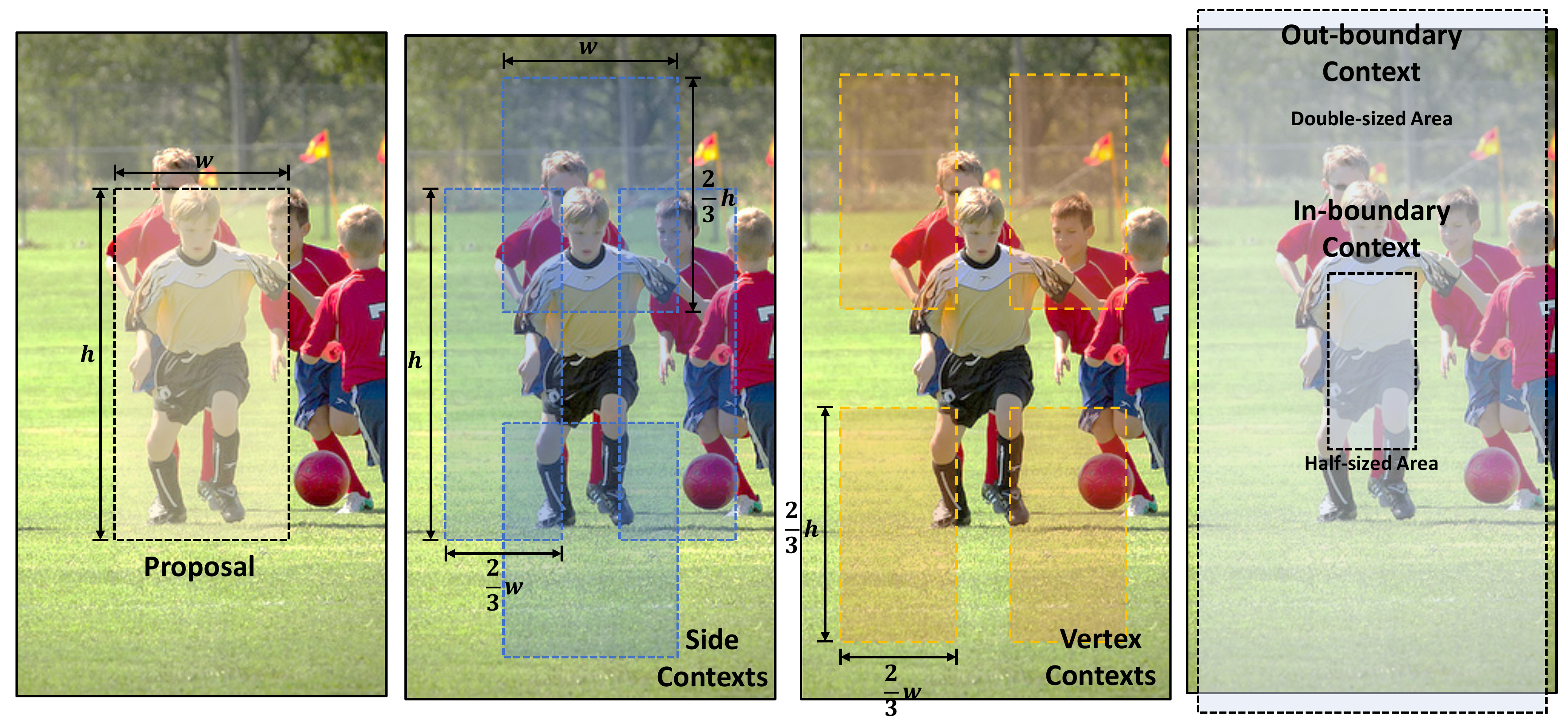}
	\caption{Three types of boundary contexts:
		(1) Side contexts represent areas centered on each side of the proposal and imply the relationship with the nearby objects and localization in the vertical and horizontal direction,
		(2) Vertex contexts represent areas centered on each vertices of the proposal and imply the relationship with the nearby objects and localization in the diagonal direction,
		(3) In and Out-boundary contexts represent the inner or outer region around the boundaries of the proposal and imply the detail or the relationship with surrounding objects.
	}
	\centering
	\label{fig:context}
\end{figure}

\subsection{Architecture}
We use a fully convolutional network that excludes the average pooling, 1000-d fully connected and softmax layers from ResNet-101~\cite{he2016deep} as backbone.
Each sub-network in BAN takes a prediction map by stacking $1\times1$ convolution from the backbone network and uses PSRoI pooling~\cite{li2016r} to calculate the objectiveness and bounding box of the given proposals.	
We employ 10 different sub-networks to deal with different boundary regions generated from $g$ for the boundary contexts.
BAN classifies and regresses a objectiveness and a bounding box of the proposal through a detection head that is an ensemble of 11 sub-networks' predictions including a sub-network for the original proposal~(Fig.~\ref{fig:overview}) .
In the learning process, each sub-network is not learned to have the same importance, but is learned to have different magnitudes of contribution
according to the sub-problems such as classification of person and relative regression of width, although it is a simple aggregation.

\subsection{Boundary Context}
We use a total of 10 boundary contexts from three different types of pre-defined boundary contexts: side, vertex and in/out-boundary context~(Fig.~\ref{fig:context}).
The RoIs for side contexts are defined as 
regions having the same height and 2/3 width of the proposal, centered at each left and right side of the proposal
and
regions having 2/3 height and same width of the proposal, centered at the other parallel sides.
The RoIs for the vertex contexts are defined as the regions having 2/3 of height and width of the proposal and are centered at each vertex of the proposal.
The RoIs for the in and out-boundary contexts are defined as a half-size region and a double-size region, respectively, sharing center point with the proposal.

\begin{figure}[!hbtp]
	\includegraphics[width=12cm]{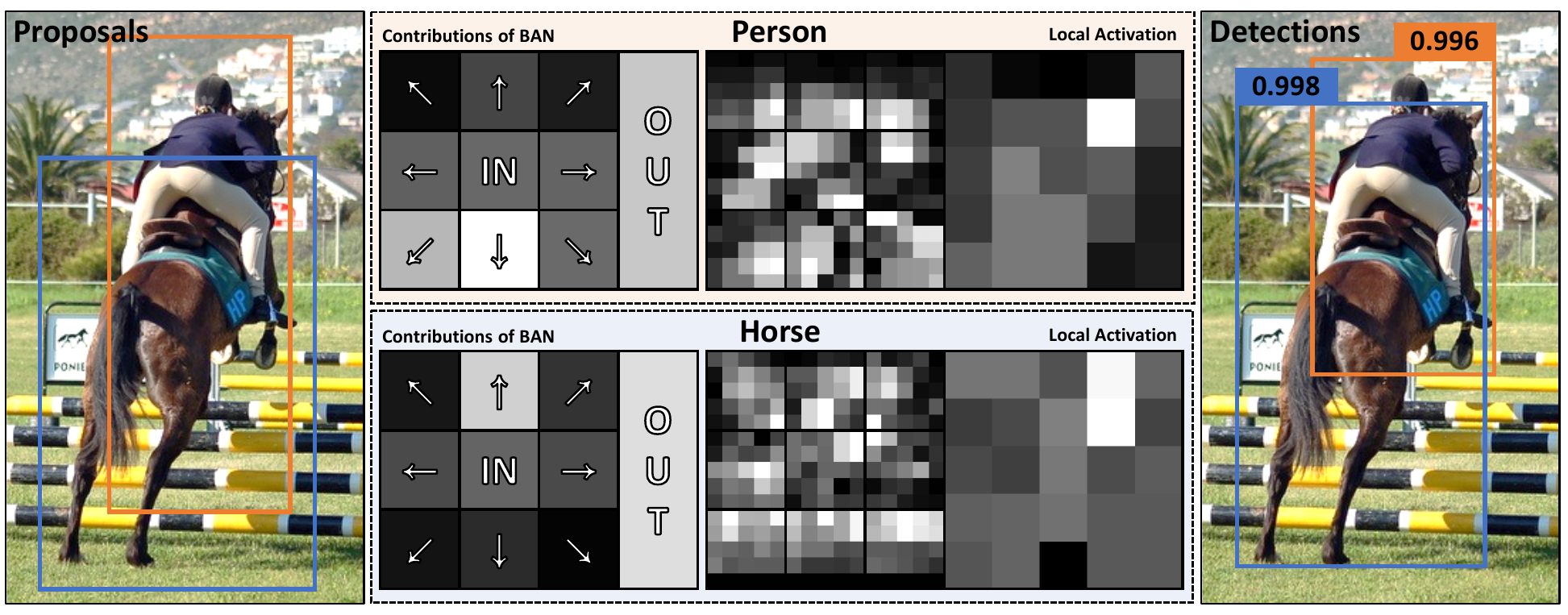}
	\caption{
		%Illustration on BAN.
		%Each feature map represents the maximum activation of the sub-networks of BAN for each boundary context.
		%Our detector classifies the proposal as cow and obtains a tightly fitted bounding box as shown on the right by adding all these results. 
		Illustration of BANs for two related objects~(person and horse) in a given image. 
		We visualize BAN with \textsf{Contribution} and \textsf{Local Activation} to show it's effectiveness more directly.
	}
	\centering
	\label{fig:cows}
\end{figure}

\begingroup
\newcolumntype{C}[1]{>{\centering\let\newline\\\arraybackslash\hspace{0pt}}m{#1}}
\begin{table}[!hbtp]
	\caption{Contribution of BAN's sub-networks for classification in PASCAL VOC.
		\texttt{Base} represents the sub-network representing the original proposal,
		each arrows represent the side and vertex context located in the corresponding direction,
		and \texttt{In} and \texttt{Out} represent the in/out-boundary contexts}
	\centering
	\scalebox{0.88}
	{
		\begin{tabular}{C{1cm}|C{1cm}|C{1cm}|C{1cm}|C{1cm}|C{1cm}|C{1cm}|C{1cm}|C{1cm}|C{1cm}|C{1cm}|C{1cm}}
			\toprule % <-- Toprule here
			&  \texttt{Base} & $\uparrow$ & $\downarrow$ & $\leftarrow$ & $\rightarrow$ &  $\nwarrow$ & $\searrow$ & $\nearrow$ & $\swarrow$  & \texttt{In} & \texttt{Out}  \\
			\midrule % <-- Midrule here
			bkgd	&	0.070	&	\textbf{0.138}	&	0.056	&	0.061	&	0.059	&	0.038	&	0.034	&	0.042	&	0.035	&	\textbf{0.328}	&	\textbf{0.140}	\\
			aero	&	0.069	&	0.112	&	0.082	&	0.103	&	0.101	&	0.056	&	0.062	&	0.047	&	0.075	&	\textbf{0.137}	&	\textbf{0.156}	\\
			bike	&	0.110	&	\textbf{0.157}	&	\textbf{0.112}	&	0.082	&	0.071	&	0.043	&	0.054	&	0.048	&	0.060	&	\textbf{0.207}	&	0.056	\\
			bird	&	0.131	&	\textbf{0.121}	&	0.094	&	0.101	&	0.095	&	0.062	&	0.054	&	0.060	&	0.055	&	\textbf{0.172}	&	0.054	\\
			boat	&	0.105	&	\textbf{0.130}	&	0.086	&	0.098	&	0.097	&	0.064	&	0.045	&	0.062	&	0.064	&	0.116	&	\textbf{0.133}	\\
			bottle	&	0.124	&	0.093	&	0.083	&	0.076	&	0.106	&	0.051	&	0.048	&	0.055	&	0.049	&	\textbf{0.227}	&	0.089	\\
			bus		&	0.074	&	\textbf{0.127}	&	0.084	&	0.089	&	0.078	&	0.054	&	0.053	&	0.053	&	0.059	&	\textbf{0.209}	&	\textbf{0.121}	\\
			car		&	0.095	&	\textbf{0.128}	&	0.081	&	0.099	&	0.111	&	0.063	&	0.071	&	0.056	&	0.075	&	\textbf{0.157}	&	0.063	\\
			cat		&	0.187	&	0.119	&	0.076	&	0.072	&	0.081	&	0.056	&	0.042	&	0.047	&	0.053	&	\textbf{0.209}	&	0.058	\\
			chair	&	0.112	&	\textbf{0.120}	&	0.115	&	0.104	&	0.105	&	0.058	&	0.057	&	0.055	&	0.060	&	\textbf{0.144}	&	0.070	\\
			cow		&	0.079	&	\textbf{0.163}	&	0.077	&	0.084	&	0.093	&	0.058	&	0.043	&	0.048	&	0.044	&	\textbf{0.255}	&	0.056	\\
			table	&	0.093	&	0.108	&	0.103	&	0.093	&	0.099	&	0.060	&	0.057	&	0.053	&	0.053	&	\textbf{0.222}	&	0.058	\\
			dog		&	0.088	&	0.108	&	0.088	&	0.079	&	0.085	&	0.060	&	0.045	&	0.049	&	0.057	&	\textbf{0.233}	&	0.107	\\
			horse	&	0.085	&	\textbf{0.137}	&	0.072	&	0.082	&	0.074	&	0.058	&	0.039	&	0.054	&	0.047	&	\textbf{0.256}	&	0.097	\\
			mbike	&	0.124	&	0.115	&	0.072	&	0.088	&	0.092	&	0.051	&	0.059	&	0.048	&	0.066	&	\textbf{0.225}	&	0.060	\\
			person	&	0.138	&	\textbf{0.148}	&	0.080	&	0.114	&	0.117	&	0.061	&	0.066	&	0.062	&	0.063	&	0.105	&	0.046	\\
			plant	&	0.096	&	\textbf{0.149}	&	0.092	&	0.095	&	0.093	&	0.059	&	0.067	&	0.050	&	0.073	&	\textbf{0.148}	&	0.078	\\
			sheep	&	0.109	&	\textbf{0.142}	&	0.112	&	0.107	&	0.118	&	0.065	&	0.053	&	0.051	&	0.059	&	\textbf{0.130}	&	0.054	\\
			sofa	&	0.107	&	\textbf{0.144}	&	0.080	&	0.088	&	0.086	&	0.068	&	0.040	&	0.066	&	0.046	&	\textbf{0.154}	&	\textbf{0.120}	\\
			train	&	0.075	&	0.097	&	0.083	&	0.090	&	0.084	&	0.058	&	0.060	&	0.055	&	0.069	&	\textbf{0.141}	&	0.190	\\
			tv		&	0.100	&	\textbf{0.134}	&	0.116	&	0.104	&	0.106	&	0.063	&	0.060	&	0.058	&	0.057	&	0.111	&	0.091	\\
			\bottomrule % <-- Bottomrule here
		\end{tabular}
	}
	\label{table:contcls}
\end{table}

\begin{table}[!hbtp]
	\caption{Contribution of BAN's sub-networks for localization in PASCAL VOC}
	\centering
	\scalebox{0.88}
	{
		\begin{tabular}{C{1cm}|C{1cm}|C{1cm}|C{1cm}|C{1cm}|C{1cm}|C{1cm}|C{1cm}|C{1cm}|C{1cm}|C{1cm}|C{1cm}}
			\toprule % <-- Toprule here
			&  \texttt{Base} & $\uparrow$ & $\downarrow$ & $\leftarrow$ & $\rightarrow$ &  $\nwarrow$ & $\searrow$ & $\nearrow$ & $\swarrow$  & \texttt{In} & \texttt{Out}  \\
			\midrule % <-- Midrule here
			$cx$	 &	0.181	&	0.088	&	0.077	&	\textbf{0.118}	&	\textbf{0.137}	&	0.058	&	0.062	&	0.053	&	0.058	&	0.077	&	0.090	\\
			$cy$	 &	0.094	&	\textbf{0.123}	&	\textbf{0.096}	&	0.046	&	0.046	&	0.037	&	0.041	&	0.033	&	0.045	&	0.055	&	\textbf{0.384}	\\
			$width$	 &	0.118	&	0.065	&	0.071	&	\textbf{0.123}	&	\textbf{0.132}	&	0.068	&	0.070	&	0.059	&	0.089	&	0.100	&	0.104	\\
			$height$ &	0.089	&	\textbf{0.131}	&	\textbf{0.105}	&	0.051	&	0.044	&	0.038	&	0.042	&	0.034	&	0.044	&	\textbf{0.212}	&	\textbf{0.209}	\\
			\bottomrule % <-- Bottomrule here
		\end{tabular}
	}
	\label{table:contloc}
\end{table}
\endgroup

\subsection{Visualization of BAN}
\label{section:sub1}

We visualize the response of feature map that is activated on the closer area to the related object~(Fig.~\ref{fig:cows}) to show the effectiveness of BAN.
\textsf{Contribution} shows that BAN is weighted more strongly to the related instance rather than backgrounds and 
\textsf{Local Activation} shows that the context is activated closer to the target. 
We also measured the classification contribution of BAN's sub-networks~(Table~\ref{table:contcls}).
The contributions are almost uniformly distributed due to large variations of the objects,
but the boundary contexts of $\uparrow$ and \texttt{In}, which can include the representative part such as head and detail, show a slightly larger contribution.
The localizations contributions demonstrate that BAN works faithfully in considering the boundary context (Table~\ref{table:contloc}).
The regression in vertical direction such as $cy$ and $height$, are highly contributed by $\uparrow$ and $\downarrow$.
In/Out-boundary contexts do not have a specific tendency but show a high contribution.
We infer that the redundancy of the regions for \texttt{base}, \texttt{in} and \texttt{out} makes them play a similar role.
We construct both visualization and contributions using PSRoI pooling based BAN for intuitive comparison.

%------------------------------------------------------------------------- 
\section{Experiments}

We conduct experiments on two different datasets of object detection and experiments with the strategies for BAN such as a combination of boundary contexts, a feature resolution of sub-networks and sharing of features.
Our BAN shows the improvement of 3.2 mAP with a threshold of 0.5 IoU from R-FCN~\cite{li2016r} and 1.2 mAP from Deformable R-FCN~\cite{dai2017deformable} on PASCAL VOC~\cite{everingham2010pascal},
and the improvement of 4.5 COCO-style mAP from R-FCN and 2.4 COCO-style mAP from Deformable R-FCN on MS COCO~\cite{lin2014microsoft}.
The experiments show that BAN improves the detection accuracy of object detection and implies that the boundary contexts has a distinct meaning for detection among each other.

\begin{figure}[!t]
	\includegraphics[width=12cm]{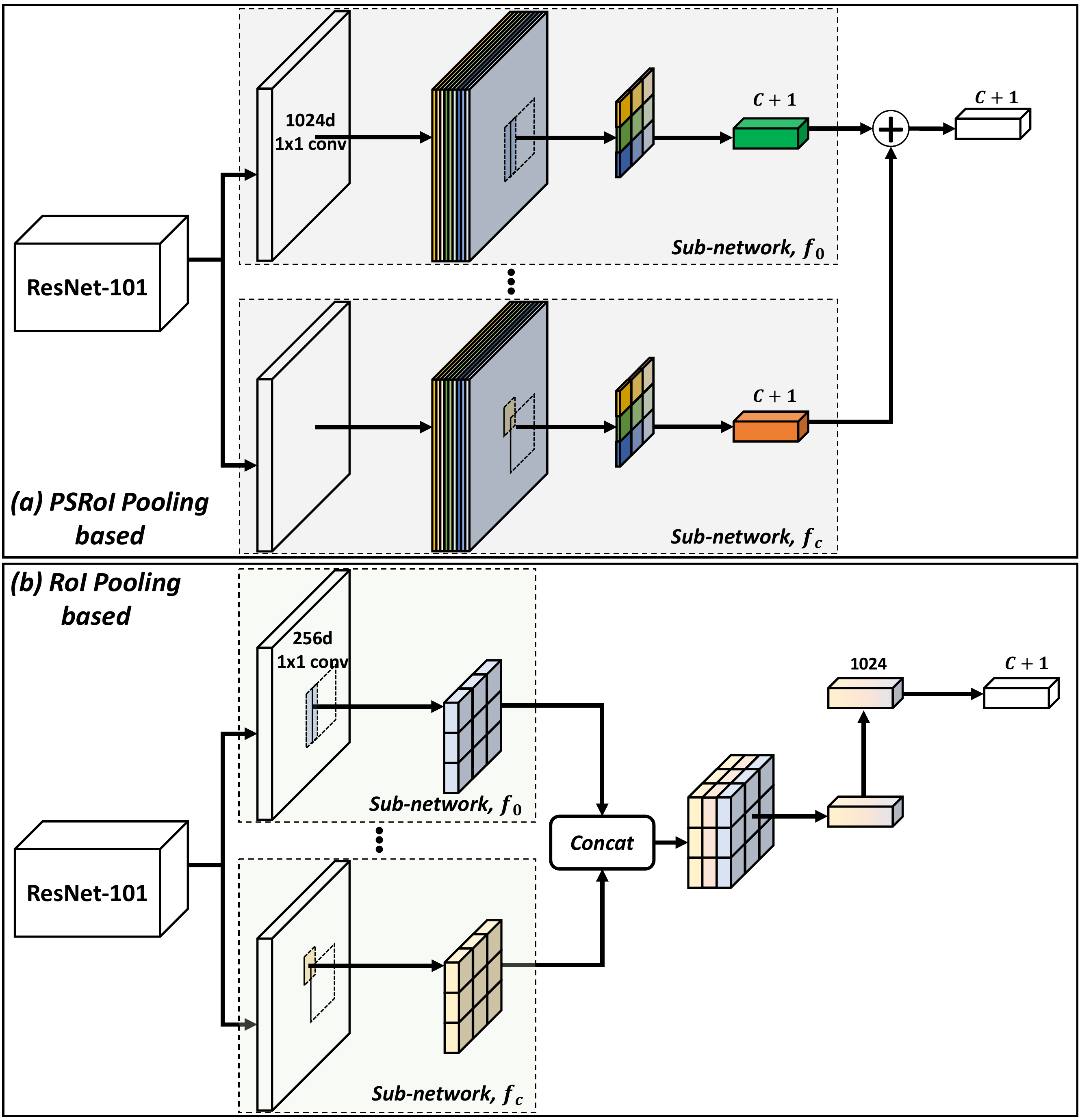}
	\caption{
		Detail structure of BAN for a classifier head of $C$ classes~(a regression head of $B$ offsets is also similarly defined). 
		The structure of BAN is empirically determined according to pooling methods~(PSRoI pooling and RoI pooling).
	}
	\centering
	\label{fig:shared}
\end{figure}

\subsection{Implementation}

\noindent\textbf{Baseline.}
We use a fully convolutional network~\cite{li2016r} that excludes the average pooling, 1000-d fully connected and softmax layers from ResNet-101~\cite{he2016deep}.	
The last convolution block~\textit{res5} in ResNet-101 has a stride of 32 pixels.
Many detection and segmentation methods employ a modified ResNet-101 that increases the receptive fields by changing the stride from 2 to 1.
To compensate this modification, the dilation is changed from 1 to 2 for all $3\times3$ convolution in the last layer. 
The last convolution block~\textit{res5} in modified ResNet-101 has a stride of 16 pixels and we use this as backbone.
We fine-tune the model from the pre-trained ResNet-101 model on ImageNet~\cite{russakovsky2015imagenet}.

\noindent\textbf{Structure.}
BAN can be implemented using any pooling methods such as RoI pooling and PSRoI pooling~(Fig.~\ref{fig:shared}).
We empirically determine the structure of BAN according to each pooling mehtod.
BAN with PSRoI pooling integrates the sub-networks that are detection heads by aggregating them.
BAN with RoI pooling extracts 256-d convolutional features from the sub-networks and builds a single detection head using the concatenated features.
Both structures improve the detection accuracy.
However, the former is easy to analyze the contributions of contexts because all detectors, including the baseline, were structurally identical, and the latter contributes to higher improvement in accuracy because it generates more distinct features for R-FCN based detectors.

\noindent\textbf{Learning.}
We use a weight decay of $0.0001$ and a momentum of $0.9$ with stochastic gradient descent~(SGD).
We train the network for 29k iterations with a learning rate of $10^{-3}$ dividing it by 10 at 20k iterations for PASCAL VOC
and for 240k iterations with a learning rate of $10^{-3}$ dividing it by 10 at 160k iterations for MS COCO.
A mini-batch consists of 2 images, which are resized such that its shorter side of image is 600 pixels.
In training, the online hard example mining~(OHEM)~\cite{shrivastava2016training} selects 128 RoIs of hard examples among 300 RoIs per image.
OHEM evaluates the multi-task loss of all proposals then discard the proposals with the small loss to make the detector more focus on difficult samples.
The detection network is trained with 4 synchronized GPUs: each GPU holds 2 images.
We use 300 RoIs per image, which is obtained from RPN and post-processed by non-maximum suppression (NMS) with a threshold of 0.3 IoU, for both learning and inference.

\noindent\textbf{Loss function.}
The loss function is defined as a sum of the classification loss and the box regression loss.
The classification loss is defined as a cross-entropy loss,
$L_{cls} (p,u)=-\log p_u$,
where $p$ is a discrete probability distribution over $K+1$ categories and $u$ is a ground-truth class.
The regression loss is defined as a smooth $L_1$ loss~\cite{girshick2015fast},
$L_{reg}(t^u,v)=\sum _{i \in \{ x,y,w,h \} } {\mathbf{smooth}_{L_1}} \left[ t_i^u-v_i \right]$,
where $t^k$ is a tuple of bounding-box regression for each of the $K$ classes, indexed by $k$, and $v$ is a tuple of a ground-truth bounding-box regression.

\noindent\textbf{Cost Analysis.}
We perform the cost analysis on the inference time and the memory consumption~(Table~\ref{tab:cost}).
The analysis is performed using ResNet-101 and RoI pooling based BAN.
Our BAN easily improves various detection methods with a reasonable increase in memory and computing time.

\begingroup
\newcolumntype{C}[1]{>{\centering\let\newline\\\arraybackslash\hspace{0pt}}m{#1}}
\begin{table}[t]
	\centering
	\begin{minipage}[t]{.6\linewidth}
		\begin{minipage}[t]{1.\linewidth}
			\centering
			\caption{Cost Analysis}
			\begin{tabular}{p{3.4cm}|C{1.4cm}|C{1.4cm}}
				\toprule % <-- Toprule here
				& \scalebox{0.7}{\textit{Inference}} & \scalebox{0.7}{$\textit{Memory}$}   \\
				& \scalebox{0.7}{\textit{Time}} & \scalebox{0.7}{$\textit{Consumption}$}   \\
				\midrule % <-- Midrule here
				\scalebox{0.8}{R-FCN~\cite{li2016r}} 					& \scalebox{0.8}{70ms}	& \scalebox{0.8}{0.8GB}   \\
				\scalebox{0.8}{R-FCN-BAN} 				& \scalebox{0.8}{97ms}	& \scalebox{0.8}{1.2GB}   \\
				\scalebox{0.8}{Deformable R-FCN~\cite{dai2017deformable}}		& \scalebox{0.8}{96ms} 	& \scalebox{0.8}{0.9GB}   \\
				\scalebox{0.8}{Deformable R-FCN-BAN}	& \scalebox{0.8}{120ms}	& \scalebox{0.8}{1.2GB}  \\
				\bottomrule % <-- Bottomrule here
			\end{tabular}
			\label{tab:cost}
		\end{minipage}%	
		\\	
		\\
		\begin{minipage}[t]{1.\linewidth}
			\centering
			\caption{Boundary Context}
			\begin{tabular}{p{3.4cm}|C{1.4cm}|C{1.4cm}}
				\toprule % <-- Toprule here
				&    \scalebox{0.8}{$\text{mAP}_{@.5}$} & \scalebox{0.8}{$\text{mAP}_{@.7}$}   \\
				\midrule % <-- Midrule here
				\scalebox{0.8}{None} 										& \scalebox{0.8}{79.54}	& \scalebox{0.8}{61.95}  \\
				\scalebox{0.8}{$\mathcal{S}$~(Sides)} 						& \scalebox{0.8}{80.23}	& \scalebox{0.8}{62.84} \\
				\scalebox{0.8}{$\mathcal{V}$~(Vertices)}					& \scalebox{0.8}{80.01}	& \scalebox{0.8}{62.13} \\
				\scalebox{0.8}{$\mathcal{B}$~(In/Out-boundary)}				& \scalebox{0.8}{79.80}	& \scalebox{0.8}{63.23} \\
				\scalebox{0.8}{$\mathcal{S}$,$\mathcal{V}$}					& \scalebox{0.8}{80.39} & \scalebox{0.8}{63.36} \\
				\scalebox{0.8}{$\mathcal{S}$,$\mathcal{V}$,$\mathcal{B}$}	& \scalebox{0.8}{80.75} & \scalebox{0.8}{64.66} \\
				\bottomrule % <-- Bottomrule here
			\end{tabular}
		\label{tab:context}
		\end{minipage}%
	\end{minipage}%
	\begin{minipage}[t]{.4\linewidth}	
		\begin{minipage}[t]{1.\linewidth}	
			\centering
			\caption{Feature Resolution}
			\begin{tabular}{p{2.0cm}|C{1cm}|C{1cm}}
				\toprule % <-- Toprule here
				&    \scalebox{0.8}{$\text{mAP}_{@.5}$} & \scalebox{0.8}{$\text{mAP}_{@.7}$}   \\
				\midrule % <-- Midrule here
				\scalebox{0.8}{- $1\times1$} 	& \scalebox{0.8}{79.39} & \scalebox{0.8}{61.36}   \\
				\scalebox{0.8}{- $3\times3$} 	& \scalebox{0.8}{80.15}	& \scalebox{0.8}{63.45}   \\
				\scalebox{0.8}{- $5\times5$}	& \scalebox{0.8}{80.75} & \scalebox{0.8}{64.66}   \\
				\scalebox{0.8}{- $7\times7$}	& \scalebox{0.8}{80.10}	& \scalebox{0.8}{63.76}   \\
				\bottomrule % <-- Bottomrule here
			\end{tabular}
		\label{tab:res}
		\end{minipage}%
		\\	
		\\
		\begin{minipage}[t]{1.0\linewidth}
			\centering
			\caption{Feature Sharing}
			\begin{tabular}{p{2cm}|C{1cm}|C{1cm}}
				\toprule % <-- Toprule here
				&    \scalebox{0.8}{$\text{mAP}_{@.5}$} & \scalebox{0.8}{$\text{mAP}_{@.7}$}   \\
				\midrule % <-- Midrule here
				\scalebox{0.8}{- Unshared}			& \scalebox{0.8}{80.05} & \scalebox{0.8}{62.80}   \\
				\scalebox{0.8}{- Shared} 				& \scalebox{0.8}{80.75} & \scalebox{0.8}{64.66}   \\
				\bottomrule % <-- Bottomrule here
			\end{tabular}
		\label{tab:share}
		\end{minipage}
		\\	
		\\
		\begin{minipage}[t]{1.0\linewidth}	
			\centering
			\caption{Pooling Method}
			\begin{tabular}{p{2cm}|C{1cm}|C{1cm}}
				\toprule % <-- Toprule here
				&    \scalebox{0.8}{$\text{mAP}_{@.5}$} & \scalebox{0.8}{$\text{mAP}_{@.7}$}   \\
				\midrule % <-- Midrule here
				\scalebox{0.8}{- PSRoI Pooling}	& \scalebox{0.8}{80.75} & \scalebox{0.8}{64.66}   \\
				\scalebox{0.8}{- ROI Pooling}	& \scalebox{0.8}{82.72} & \scalebox{0.8}{67.84}   \\
				\bottomrule % <-- Bottomrule here
			\end{tabular}
		\label{tab:pool}
		\end{minipage}%	
	\end{minipage}%	
\end{table}
\endgroup

\subsection{Comparison with Strategies for BAN}

We experiment the strategies for BAN such as different combinations of boundary contexts,
various feature resolutions of sub-networks, feature sharing and pooling method
to construct effective BAN.
The experiments are performed using ResNet-101 and PSRoI pooling based BAN on PASCAL VOC.

\noindent\textbf{Boundary Context.}	
We conduct experiments on the types of boundary contexts~(side, vertex and in/out-boundary contexts) and the combinations of the types~(Table~\ref{tab:context}).
All boundary contexts shows the meaningful improvement in the detection accuracy
and the combination of the all three types of boundary contexts improves $\text{mAP}_{@0.5}$ by 1.21 and $\text{mAP}_{@0.7}$ by 2.71.
This experiment shows that each boundary context have a distinct meaning for detection.

\noindent\textbf{Feature Resolution.}
We conduct experiments on the feature resolution~$k \times k$ of sub-networks from $1\times1$ to $7\times7$~(Table~\ref{tab:res}).
The feature resolution of $5\times5$ shows the highest improvement and $1\times1$ degrades the detection accuracy as it crushes the boundary contexts.

\noindent\textbf{Feature Sharing.}
Each sub-network consists of a $1024$ dimensional $1\times1$ convolution and the following relu for feature extraction
and a $(C+1)k^2$ dimensional $1\times1$ convolution as classification and a $8k^2$ dimensional $1\times1$ convolution as regression for detection heads~(Table~\ref{tab:share}).
The different use of $1 \times 1$ convolution for feature extraction lead to the improvement of 0.73 point in mAP.
This experiment implies that the boundary context transfers a distinctive influence to the feature level as well as the detection head in learning.

\noindent\textbf{Pooling.}
The implementation of BAN is slightly different depending on the pooling method for extracting the visual context.
We conduct experiments on two pooling methods: RoI pooling and PSRoI pooling.
PSRoI pooling requires a small amount of resources because it is fully convolutional.
RoI pooling highly improves the accuracy in detection because it is easy to extract the fundamental convolutional features for the boundary context~(Table~\ref{tab:pool}).

\begingroup
\newcolumntype{C}[1]{>{\centering\let\newline\\\arraybackslash\hspace{0pt}}m{#1}}
\begin{table}[t]
	\caption{Evaluation on PASCAL VOC 2007 and MS COCO \texttt{test-dev}}
	\begin{center}
		\begin{tabular}{p{3.3cm}|C{1.1cm}|C{1.1cm}|C{1.1cm}|C{1.1cm}|C{1.1cm}|C{1.1cm}|C{1.1cm}}
			\toprule % <-- Toprule here
			& \multicolumn{2}{c|}{\scalebox{0.9}{VOC 2007}} & \multicolumn{5}{c}{\scalebox{0.9}{MS COCO test-dev}} \\
			& \scalebox{0.9}{$\mathrm{mAP_{@.5}}$} & \scalebox{0.9}{$\mathrm{mAP_{@.7}}$} 
			& \scalebox{0.9}{$\mathrm{mAP}$} & \scalebox{0.9}{$\mathrm{mAP_{@.5}}$} & 
			\scalebox{0.9}{$\mathrm{mAP_{@S}}$} & \scalebox{0.9}{$\mathrm{mAP_{@M}}$} & \scalebox{0.9}{$\mathrm{mAP_{@L}}$}  \\
			\midrule % <-- Midrule here
			\scalebox{0.8}{Faster RCNN~\cite{ren2015faster}}& 76.4 & - & 30.3 & 52.1 &  9.9 & 32.2 & 47.4 \\
			\scalebox{0.8}{YOLOv2~\cite{redmon2017yolo9000}}& 79.5 & - & 21.6 & 44.0 &  5.0 & 22.4 & 35.5 \\
			\scalebox{0.8}{SSD513~\cite{liu2016ssd}}		& 76.8 & - & 31.2 & 50.4 & 10.2 & 34.5 & 49.8 \\
			\scalebox{0.8}{R-FCN~\cite{li2016r}}			& 79.5 & 62.0 & 29.9 & 50.8 & 11.0 & 32.2 & 43.9 \\
			\scalebox{0.8}{R-FCN-BAN}					 	& 82.7 & 67.8 & 34.4 & 58.5 & 17.8 & 37.7 & 46.0 \\
			\scalebox{0.8}{Deformable R-FCN~\cite{dai2017deformable}}		& 82.2 & 67.6 & 34.5 & 55.0 & 14.0 & 37.7 & 50.3 \\
			\scalebox{0.8}{Deformable R-FCN-BAN}			& 83.4 & 70.0 & 36.9 & 58.5 & 15.8 & 40.0 & 53.6 \\
			\bottomrule % <-- Bottomrule here
		\end{tabular}
	\end{center}
	\label{tab:result}
\end{table}
\endgroup

\subsection{Experiments on PASCAL VOC}

\begin{figure}[!htbp]
	\begin{center}
		\subfigure{\includegraphics[width=12.2cm]{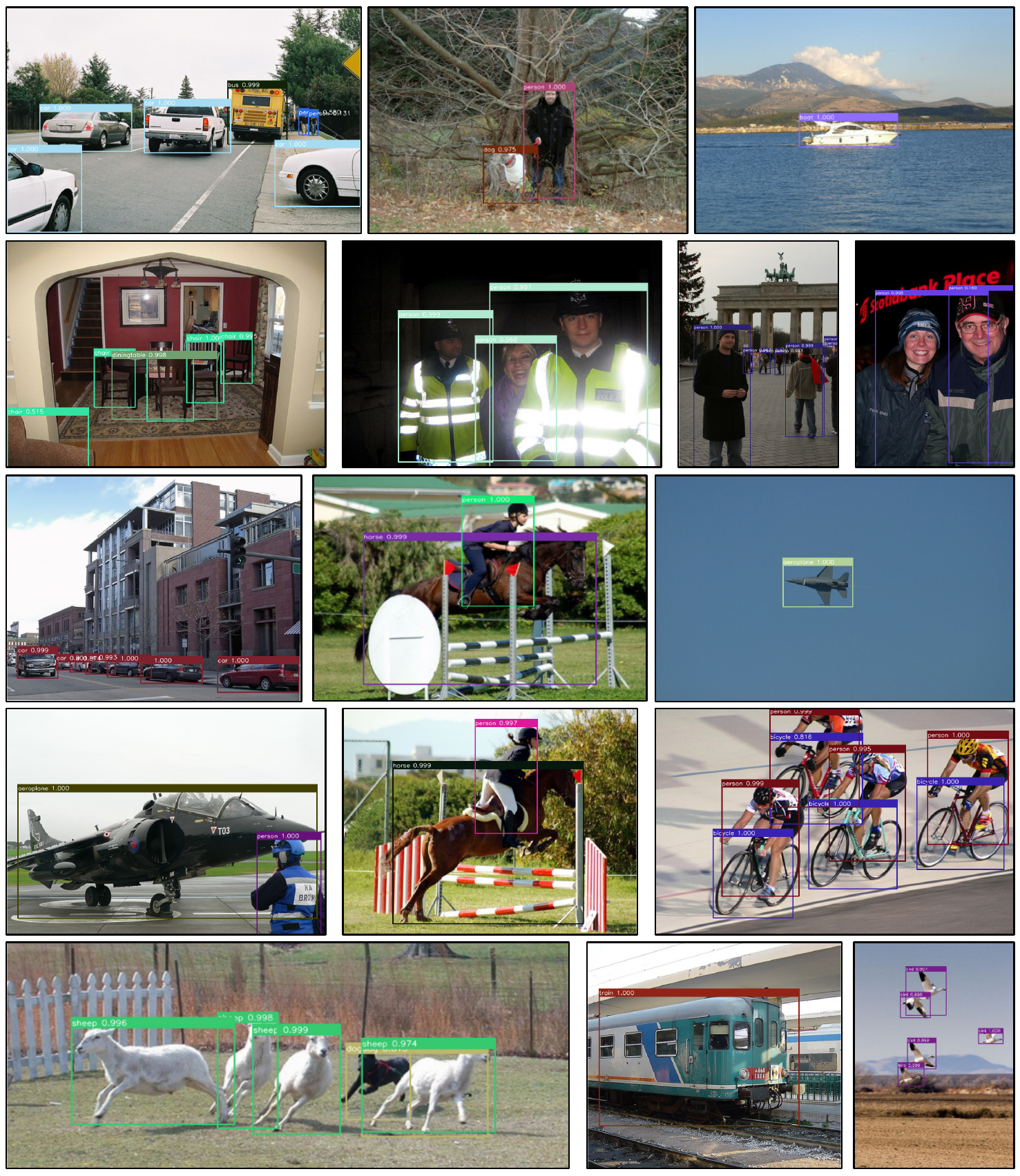}}
		\caption{
			Examples of object detection results on PASCAL VOC 2007 test set using our method~(83.4\% mAP).
			The network is based on ResNet-101 and the training data is 07+12 trainval.}
		\label{fig:result}
	\end{center}
	
	\captionof{table}{Detailed detection results on PASCAL VOC 2007 test set}
	\begin{center}
		\scalebox{0.70}
		{
			\begin{tabular}
				{l|c| >{\centering}m{0.6cm}>{\centering}m{0.6cm}>{\centering}m{0.6cm}>{\centering}m{0.6cm}>{\centering}m{0.6cm}>{\centering}m{0.6cm}>{\centering}m{0.6cm}>{\centering}m{0.6cm}>{\centering}m{0.6cm}>{\centering}m{0.6cm}
					>{\centering}m{0.6cm}>{\centering}m{0.6cm}>{\centering}m{0.6cm}>{\centering}m{0.6cm}>{\centering}m{0.8cm}>{\centering}m{0.6cm}>{\centering}m{0.6cm}>{\centering}m{0.6cm}>{\centering}m{0.6cm}c}
				\toprule % <-- Toprule here
				Method &  mAP
				& \scalebox{0.8}{aero} & \scalebox{0.8}{bike} & \scalebox{0.8}{bird} & \scalebox{0.8}{boat}
				& \scalebox{0.8}{bottle} & \scalebox{0.8}{bus} & \scalebox{0.8}{car} & \scalebox{0.8}{cat} & \scalebox{0.8}{chair} & \scalebox{0.8}{cow} 
				& \scalebox{0.8}{table} & \scalebox{0.8}{dog} & \scalebox{0.8}{horse} & \scalebox{0.8}{mbike} 
				& \scalebox{0.8}{person} & \scalebox{0.8}{plant} & \scalebox{0.8}{sheep} & \scalebox{0.8}{sofa} & \scalebox{0.8}{train} & \scalebox{0.8}{tv}  \\ 
				\midrule % <-- Midrule here
				Faster R-CNN &  \scalebox{0.8}{76.4}
				& \scalebox{0.8}{79.8} & \scalebox{0.8}{80.7} & \scalebox{0.8}{76.2} & \scalebox{0.8}{68.3} 
				& \scalebox{0.8}{55.9} & \scalebox{0.8}{85.1} & \scalebox{0.8}{85.3} & \scalebox{0.8}{89.8}
				& \scalebox{0.8}{56.7} & \scalebox{0.8}{87.8}  
				& \scalebox{0.8}{69.4} & \scalebox{0.8}{88.3} & \scalebox{0.8}{88.9} & \scalebox{0.8}{80.9} 
				& \scalebox{0.8}{78.4} & \scalebox{0.8}{41.7} & \scalebox{0.8}{78.6} & \scalebox{0.8}{79.8}
				& \scalebox{0.8}{85.3} & \scalebox{0.8}{72.0}  \\
				R-FCN~\cite{li2016r} & \scalebox{0.8}{79.5}
				& \scalebox{0.8}{82.5} & \scalebox{0.8}{83.7} & \scalebox{0.8}{80.3} & \scalebox{0.8}{69.0} 
				& \scalebox{0.8}{69.2} & \scalebox{0.8}{87.5} & \scalebox{0.8}{88.4} & \scalebox{0.8}{65.4}
				& \scalebox{0.8}{65.4} & \scalebox{0.8}{87.3}  
				& \scalebox{0.8}{72.1} & \scalebox{0.8}{87.9} & \scalebox{0.8}{88.3} & \scalebox{0.8}{81.3} 
				& \scalebox{0.8}{79.8} & \scalebox{0.8}{54.1} & \scalebox{0.8}{79.6} & \scalebox{0.8}{78.8}
				& \scalebox{0.8}{87.1} & \scalebox{0.8}{79.5}  \\
				R-FCN-BAN & \scalebox{0.8}{82.7}
				& \scalebox{0.8}{89.1} & \scalebox{0.8}{88.4} & \scalebox{0.8}{80.7} & \scalebox{0.8}{76.9} 
				& \scalebox{0.8}{73.3} & \scalebox{0.8}{89.6} & \scalebox{0.8}{88.8} & \scalebox{0.8}{89.5}
				& \scalebox{0.8}{69.5} & \scalebox{0.8}{88.0}  
				& \scalebox{0.8}{74.5} & \scalebox{0.8}{90.0} & \scalebox{0.8}{89.3} & \scalebox{0.8}{86.8} 
				& \scalebox{0.8}{80.5} & \scalebox{0.8}{57.6} & \scalebox{0.8}{84.3} & \scalebox{0.8}{84.7}
				& \scalebox{0.8}{88.5} & \scalebox{0.8}{84.5}  \\
				DR-FCN~\cite{dai2017deformable} & \scalebox{0.8}{82.2}
				& \scalebox{0.8}{85.9} & \scalebox{0.8}{89.3} & \scalebox{0.8}{80.7} & \scalebox{0.8}{74.8} 
				& \scalebox{0.8}{72.4} & \scalebox{0.8}{88.2} & \scalebox{0.8}{88.8} & \scalebox{0.8}{89.5}
				& \scalebox{0.8}{69.0} & \scalebox{0.8}{88.2}  
				& \scalebox{0.8}{75.4} & \scalebox{0.8}{89.7} & \scalebox{0.8}{89.4} & \scalebox{0.8}{84.5} 
				& \scalebox{0.8}{83.4} & \scalebox{0.8}{57.3} & \scalebox{0.8}{84.9} & \scalebox{0.8}{82.3}
				& \scalebox{0.8}{87.6} & \scalebox{0.8}{82.7}  \\
				DR-FCN-BAN & \scalebox{0.8}{83.4}
				& \scalebox{0.8}{88.0} & \scalebox{0.8}{89.5} & \scalebox{0.8}{80.6} & \scalebox{0.8}{77.0} 
				& \scalebox{0.8}{73.4} & \scalebox{0.8}{88.8} & \scalebox{0.8}{89.0} & \scalebox{0.8}{89.8}
				& \scalebox{0.8}{70.7} & \scalebox{0.8}{88.4}  
				& \scalebox{0.8}{77.3} & \scalebox{0.8}{90.2} & \scalebox{0.8}{89.4} & \scalebox{0.8}{87.5} 
				& \scalebox{0.8}{84.6} & \scalebox{0.8}{58.2} & \scalebox{0.8}{85.6} & \scalebox{0.8}{85.3}
				& \scalebox{0.8}{88.2} & \scalebox{0.8}{85.9}  \\
				\bottomrule % <-- Bottomrule here
			\end{tabular}
		}
		\label{tab:rest}
	\end{center}
\end{figure}

We evaluate the proposed BAN on PASCAL VOC~\cite{everingham2010pascal} that has 20 object categories (Fig.~\ref{fig:result}).
We train the models on the union set of VOC 2007 and VOC 2012 trainval,~\texttt{07+12}, (16,551 images),
and evaluate on VOC 2007 \texttt{test} set (4,952 images).
Detection accuracy is measured by mean Average Precision (mAP).
BAN improves 3.2 mAP with a threshold of 0.5 IoU and 5.8 mAP with a threshold of 0.7 from R-FCN~\cite{li2016r} and 1.2 mAP with a threshold of 0.5 IoU and 2.4 mAP with a threshold of 0.7 from Deformable R-FCN~\cite{dai2017deformable}~(Table~\ref{tab:result} and~\ref{tab:rest}).

\subsection{Experiments on MS COCO}

We evaluate the proposed BAN on MS COCO dataset~\cite{lin2014microsoft} that has 80 object categories.
We train the models on the union set of 80k training set and 40k validation set (\texttt{trainval}),
and evaluate on 20k \texttt{test-dev} set.
The COCO-style metric denotes mAP, which is the average AP across thresholds of IoU from 0.5 to 0.95 with an interval of 0.05.
Our BAN improves 4.5 COCO-style mAP and 7.7 mAP with a threshold of 0.5 IoU from R-FCN~\cite{li2016r} and 2.4 COCO-style mAP and 3.5 mAP with a threshold of 0.5 IoU from Deformable R-FCN~\cite{dai2017deformable}~(Table~\ref{tab:result}).
We obtain the higher improvement in the detection accuracy for MS COCO with various classes and challenging environments, than PASCAL VOC.

%------------------------------------------------------------------------- 
\section{Conclusions}

We propose a boundary aware network~(BAN) designed to exploit the boundary contexts 
and study empirically the influence of the boundary context on classification and bounding box regression.
To show the effectiveness of BAN, we visualize the activation of the sub-networks according to the boundary contexts and 
empirically show that the boundary contexts of BAN contributes more strongly to the detection head intuitively related to the boundary context.
These related contribution suggests that BAN implies distinct meanings than naive ensemble of sub-networks.	
We evaluate our method on PASCAL VOC detection benchmark dataset, which has 20 object categories and MS COCO dataset, which has 80 object categories.
Our BAN improves mAP by 3.2 point from R-FCN and 1.2 point from Deformable R-FCN on PASCAL VOC and improves the COCO-style mAP by 4.5 point from R-FCN and 2.4 point from Deformable R-FCN on MS COCO.
BAN allows the convolution network to provide an additional source of contexts for detection and selectively focus on more important contexts,
and it can be generally applied to many other detection method as well to enhance the accuracy in detection.

As a future study, we will improve the detection accuracy by applying BAN to the entire network including RPN.
In addition, we plan to develop a general version of BAN based on this study of the influence and relationship among the boundary contexts.

\section*{Acknowledgments}

This work was supported by IITP grant funded by the Korea government (MSIT)
(IITP-2014-3-00059, Development of Predictive Visual Intelligence Technology, 
IITP-2017-0-00897, SW Starlab support program,
and 
IITP-2018-0-01290, Development of Open Informal Dataset and Dynamic Object Recognition Technology Affecting Autonomous Driving)

\clearpage

\bibliographystyle{splncs04}
\bibliography{egbib}

\end{document}